\documentclass{article}

\usepackage{PRIMEarxiv}

\usepackage[utf8]{inputenc} 
\usepackage[T1]{fontenc}    
\usepackage{hyperref}       
\usepackage{url}            
\usepackage{booktabs}       
\usepackage{amsfonts}       
\usepackage{nicefrac}       
\usepackage{microtype}      
\usepackage{lipsum}
\usepackage{fancyhdr}       
\usepackage{graphicx}       
\graphicspath{{media/}}     

\usepackage{xcolor,colortbl}
\usepackage{pdfpages}
\usepackage{stackengine,collcell,makecell}

\usepackage[resetlabels,labeled]{multibib}
\usepackage{tcolorbox}

\newcites{A}{Appendix References}

\title{DC-Check: A Data-Centric AI checklist to guide the development of reliable machine learning systems}

\author{
  Nabeel Seedat \\
  University of Cambridge\\
 \texttt{ns741@cam.ac.uk}
   \And
  Fergus Imrie \\
  UCLA \\
  \texttt{imrie@g.ucla.edu} \\
   \And
  Mihaela van der Schaar \\
  University of Cambridge \\
  Alan Turing Institute \\
  \texttt{mv472@cam.ac.uk} \\
}

\begin{document}
\maketitle

\begin{abstract}
  While there have been a number of remarkable breakthroughs in machine learning (ML), much of the focus has been placed on model development.  However, to truly realize the potential of machine learning in real-world settings, additional aspects must be considered across the ML pipeline. Data-centric AI is emerging as a unifying paradigm that could enable such reliable end-to-end pipelines. However, this remains a nascent area with no standardized framework to guide practitioners to the necessary data-centric considerations or to communicate the design of data-centric driven ML systems. To address this gap, we propose DC-Check, an actionable checklist-style framework to elicit data-centric considerations at different stages of the ML pipeline: Data, Training, Testing, and Deployment. This data-centric lens on development aims to promote thoughtfulness and transparency prior to system development. Additionally, we highlight specific data-centric AI challenges and research opportunities. DC-Check is aimed at both practitioners and researchers to guide day-to-day development. As such, to easily engage with and use DC-Check and associated resources, we provide a \href{https://www.vanderschaar-lab.com/dc-check/}{DC-Check companion website} (https://www.vanderschaar-lab.com/dc-check/). The website will also serve as an updated resource as methods and tooling evolve over time.

\end{abstract}

\keywords{data-centric AI, ML pipelines, reliable ML, checklist, data, training, deployment, testing}

\section{Introduction}
Machine learning has seen numerous algorithmic breakthroughs in recent years. This has increased the appetite for adoption across a variety of industries: from recommendation systems in e-Commerce \cite{zhao2019study}, speech recognition for virtual assistants \cite{swarup2019improving}, natural language processing for machine translation \cite{stahlberg2020neural},  computer vision for medical image analysis \cite{esteva2021deep}, and a deluge of tabular data applications across healthcare \cite{alaa2021machine}, finance \cite{moscato2021benchmark}, and manufacturing \cite{shwartz2022tabular}. 
In realizing these leaps, the focus in the ML community has often been to optimize algorithms for state-of-the-art (SOTA) performance on benchmark datasets, whilst neglecting a swathe of considerations necessary in real-world settings including, but not limited to: data curation, data quality assessment and cleaning, characterizing subgroups in data, model robustness, monitoring and accounting for data drifts, and much more. This gap to reality has meant that real-world systems have, with notable exceptions, been unable to unequivocally and consistently perform as expected.

In fact, there have been numerous high-profile ML failures such as gender and racial biases in Twitter's image cropping algorithm \cite{chowdhury}, football object tracking algorithms latching onto spurious correlations and mistaking the linesman's head for the ball \cite{vincent_2020}, or Google Health's diabetic retinopathy system, which failed on images with lower quality than development \cite{beede2020human}. Furthermore, approximately 85\% of industrial AI/ML systems are projected to ``deliver erroneous outcomes due to bias in data or algorithms’’, while only 53\% bridge the gap between prototypes and production \cite{costello}. This highlights that simply deploying a seemingly highly predictive model is but one piece of the puzzle for a reliable ML system in practice. In fact, a recurring theme across anecdotes is the \emph{data} being a driver of these failures.

The ML community has attempted to improve the usage of ML systems with \textit{Machine Learning Operations (MLOps)} \cite{ symeonidis2022mlops, makinen2021needs,john2021towards}, \textit{Robust Machine Learning} \cite{song2020learning,northcutt2021confident, rahimian2019distributionally, hullermeier2021aleatoric} and \textit{Trustworthy Machine learning} \cite{varshney2019trustworthy,chouldechova2020snapshot,du2019techniques}. 
While individually each addresses important yet different challenges, in isolation, none are sufficient as a  ``silver bullet'' for reliable ML at the \emph{end-to-end ML pipeline level}, i.e. covering the ML lifecycle from data all the way to deployment. Such considerations are especially relevant as ML systems are increasingly adopted in diverse, high-stakes, and complex domains, such as healthcare, finance, law enforcement, etc. Consequently, we believe that a systematic and principled framework that can be used to guide the development of end-to-end ML pipelines is sorely needed. 

\begin{figure}[t]
    \centering
    \includegraphics[width=0.85\columnwidth]{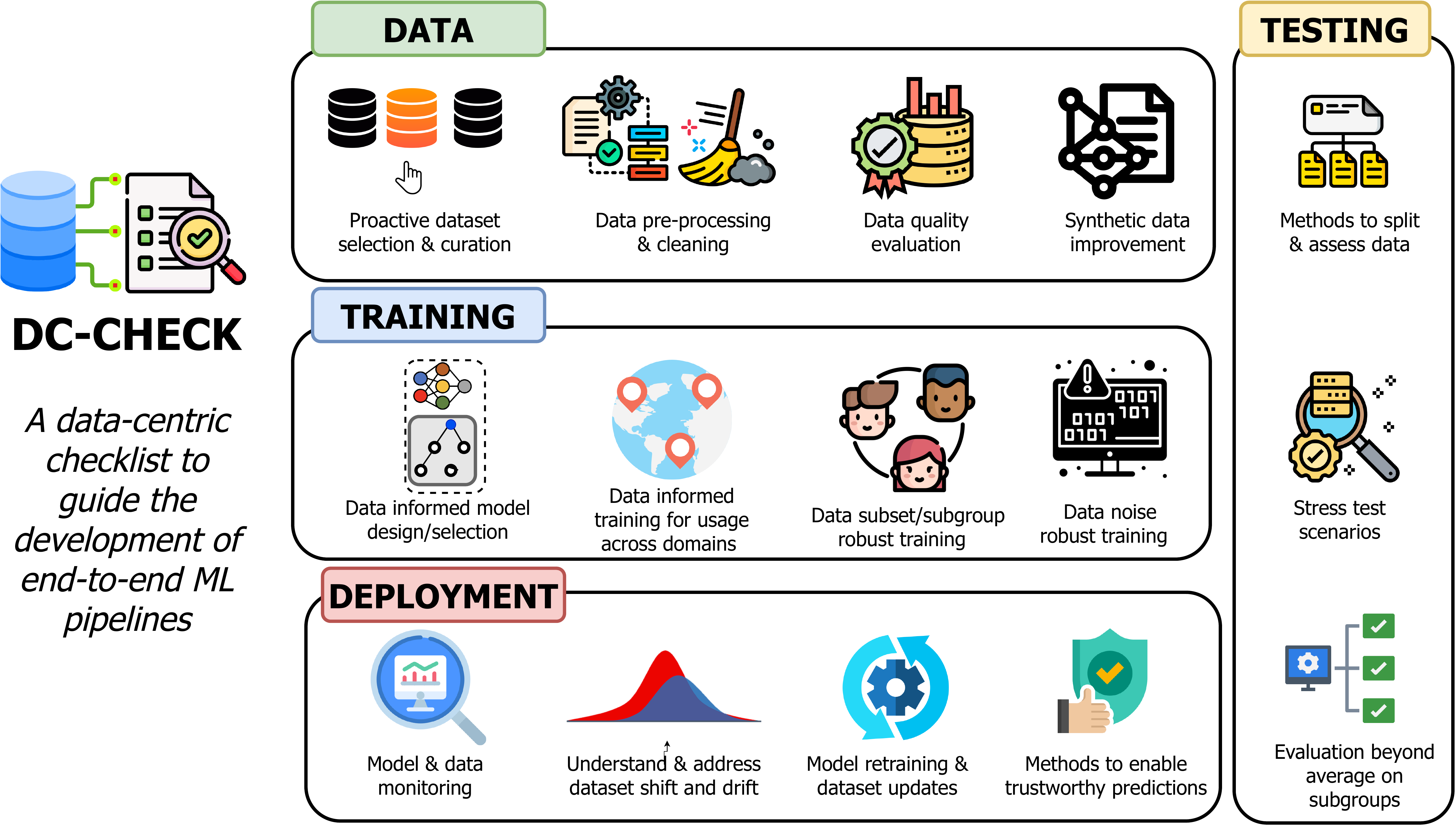}
    \caption{DC-Check: Detailed components with a data-centric lens considered across the pipeline.}
    \label{fig:simple}
\end{figure}

Specifically, we propose the nascent area of \textbf{``data-centric AI''} as a harmonizing umbrella. 
Data plays a critical role in ML, with its characteristics and quality influencing each stage of the pipeline. 
Consequently, our vision is a systematic and principled framework that places ``data at the center''. Real-world data is often diverse, noisy and ever-changing, thus we believe adopting a data-centric lens at each stage of the pipeline is crucial to developing reliable systems. 

At this point, it is pertinent to define and contrast current notions of \emph{data-centric} AI  vs \emph{model-centric} AI (the current de facto standard).  \emph{Data-centric} AI views model or algorithmic refinement as less important (and in certain settings algorithmic development is even considered as a solved problem), and instead seeks to systematically improve the data used by ML systems. Conversely, in \emph{model-centric} AI, the data is considered an asset adjacent to the model and is often fixed or static (e.g. benchmarks), whilst improvements are sought specifically to the model or algorithm itself. We believe that the current focus on models and architectures as a panacea in the ML community is often a source of brittleness in real-world applications. In this work, we outline why the data work, often undervalued as merely operational, is key to unlocking reliable ML systems in the wild.

That said, in the current form, definitions of data-centric vs model-centric are insufficient when considering end-to-end pipelines. A limitation is the binary delineation of working only on the data or model. In this work, we go further and call for an expanded definition of \emph{data-centric AI} such that a data-centric lens is applicable for end-to-end pipelines.  

\begin{tcolorbox}[width=\textwidth,colback={blue!5},title={DEFINITION: DATA-CENTRIC AI},colframe=blue!75!black]

\emph{Data-centric AI encompasses methods and tools to systematically characterize, evaluate, and monitor the underlying data used to train and evaluate models}.
At the ML pipeline level, this means that the considerations at each stage should be informed in a data-driven manner. We term this a \textbf{data-centric lens}. Since data is the fuel for any ML system, we should keep a sharp focus on the data, yet rather than ignoring the model, we should leverage the data-driven insights as feedback to systematically improve the model. 
\end{tcolorbox}

\begin{figure}[t]
    \centering
    \includegraphics[width=0.85\columnwidth]{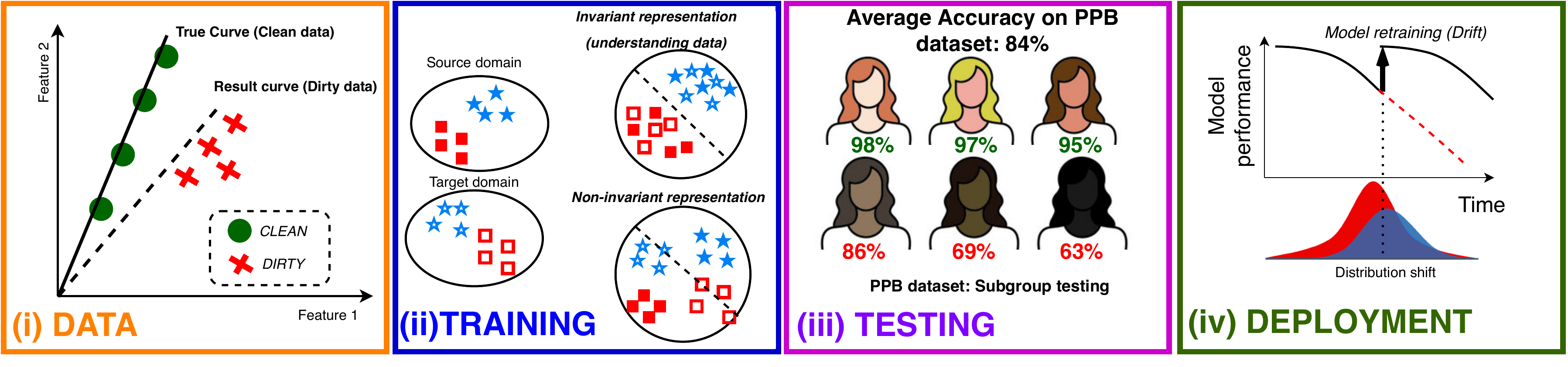}
    \caption{Examples highlighting the importance of each DC-Check component. (i) Data: Without consideration of data cleaning, models fit on  ``dirty'' data  are suboptimal model (see \cite{krishnan2016activeclean}), (ii) Training: Without consideration of data one might train models with representations which do not generalize (see \cite{csurka2017domain,li2018domain}), (iii)Testing: without consideration of testing on subgroups, one could miss underperformance (see \cite{dulhanty2019auditing}), (iv) Deployment: without consideration of data drift, model performance degrades over time (see \cite{vela2022temporal,bayram2022concept}).}
    \vspace{-0.3cm}
    \label{fig:uses}
\end{figure}

Despite being a nascent area, data-centric AI has been raised as an important concept to improve ML systems \cite{ng2021,liang2022advances, polyzotis2021can,seedat2022data}. However, currently there is no standardized process in which to communicate the design of data-centric ML pipelines. More specifically, there is no guide to the necessary considerations for data-centric AI systems, making the agenda hard to practically engage with.
To address this gap, we propose \textbf{DC-Check}, an actionable checklist that advocates for a \emph{data-centric lens} encompassing the following stages of the ML pipeline (see Figure \ref{fig:simple}): 
\begin{itemize}
    \item \textbf{Data}: Considerations to improve the quality of data used for model training, such as proactive data selection, data curation, and data cleaning.
    \item \textbf{Training}: Considerations based on understanding the data to improve model training, such as data informed model design, domain adaptation, and robust training.
    \item \textbf{Testing}: Considerations around novel data-centric methods to test ML models, such as informed data splits, targeted metrics and stress tests and evaluation on subgroups.
    \item \textbf{Deployment}: Considerations based on data post-deployment, such as data and model monitoring, model adaptation and retraining, and uncertainty quantification.

\end{itemize}


The ML community has begun to engage with similar concepts of documentation for ML processes. In particular, Datasheets for Datasets \cite{gebru2021datasheets} 
proposes documentation guidelines for new datasets, while Model Cards for Model Reporting \cite{mitchell2019model} advocates for standardized reporting of models, including training data, performance measures, and limitations.
We consider these documentation paradigms as complementary to DC-Check, as first they both only consider one specific part of the pipeline (data and model, respectively), and second, their focus is on improving documentation \textit{after} the fact.  In contrast, DC-Check covers the entire pipeline and further we advocate for engaging with these considerations \textit{before} building ML systems. We hope that DC-Check and the data-centric lens will encourage thoughtful data-centric decision-making from the offset regarding the impact of data and its influence on the design of ML systems at each stage of the ML pipeline. In Figure \ref{fig:uses} we showcase examples at different pipeline stages, where failure to consider the data-centric issue has resulted in failures.

DC-Check is aimed at both practitioners (ML engineers, data scientists, software engineers) and researchers. Each component of DC-Check includes a set of data-centric questions to guide users, 
thereby helping developers to clearly understand potential challenges. From the practical side, we also suggest concrete data-centric tools and modeling approaches based on these considerations. In addition to the checklist that guides projects, we also include research opportunities necessary to advance the nascent research area of data-centric AI. 

Going beyond a documentation tool, DC-Check supports practitioners and researchers in achieving greater transparency and accountability with regard to data-centric considerations for ML pipelines.  We believe that this type of transparency and accountability provided by DC-Check can be useful to policymakers, regulators, and organization decision makers to understand the design considerations at each stage of the ML pipeline.

To guide usage, 
we provide worked examples of DC-Check for two real-world projects the in Appendix, along with example tooling for each component of DC-Check (Tables 2-5). Finally, before delving into DC-Check, we highlight that the relevant data-centric considerations will naturally differ based on the ``type'' and ``level'' of reliability required. In particular, different applications have different stakes and this will inform which aspects of DC-Check are most relevant. As a consequence, the DC-Check questions are by no means prescriptive, complete, or exhaustive. We expect that certain questions will be more relevant depending on the existing workflows, context, stakeholders, or use cases.

Of course, we expect that the data-centric field will develop over time as the opportunities included in this paper and beyond are addressed and new tools are developed. Consequently, while we believe that DC-Check covers fundamental data-centric design considerations, the tooling and methods for carrying out the tasks will naturally evolve. Hence, in addition to these fundamental considerations contained in the checklist, we couple the paper with a \href{https://www.vanderschaar-lab.com/dc-check/}{DC-Check companion website}(https://www.vanderschaar-lab.com/dc-check/). The website will serve as an easy-to-use gateway to use and engage with DC-Check. It will also keep track of method and tooling changes and subsequently be updated over time. We envision that this companion website will serve as a data-centric community hub, allowing researchers and practitioners to engage with and contribute to DC-Check.

\section{\textbf{Data}: a data-centric lens informing dataset curation}

\begin{figure}[!h]
    \centering
    \includegraphics[width=0.9\columnwidth]{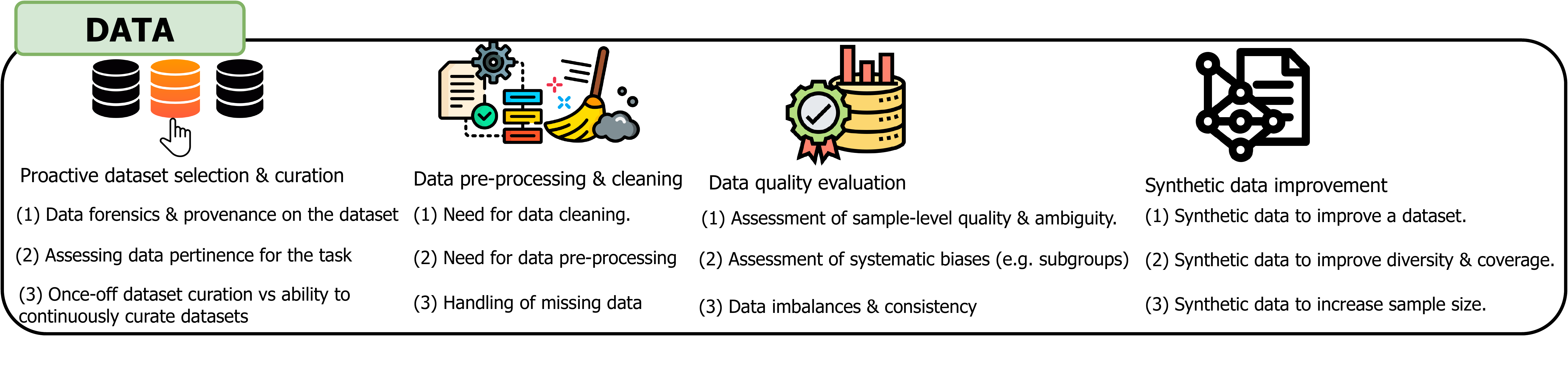}
    \caption{DATA: considerations for more systematic data-centric curation of datasets}
    \label{fig:data}
\end{figure}

The performance and robustness of ML systems often depends on the characteristics and quality of data \cite{jain2020overview,renggli2021data,Sambasivan}. In particular, data with imperfections or limitations can lead to suboptimal model performance. Consequently, systematic data curation is crucial to ensure ML reliability.  Such imperfections could arise for reasons including, but not limited to: 
\begin{itemize}
    \item Poor quality data: data is noisy, corrupted or labels are ambiguous or mislabelled.
    \item Biased data: collection process introduces bias or specific groups have coverage imbalances.
    \item Missing data: specific features are missing either at random or not at random.
\end{itemize}

Such data issues result in so-called ``Data Cascades'' \cite{Sambasivan}, where problems with the data affect downstream tasks. In mitigating this, the ``data work'' usually begins with exploratory data analysis (EDA). Based on EDA, next steps involve data cleaning and curation, which involves a large amount of human effort, often performed in an ad hoc manner. Not only is the approach prone to human variability, but dataset curation and preparation is also tedious and time consuming, typically accounting for 80\% of the time in a data science project \cite{li2021cleanml,krishnan2019alphaclean}. 
Hence, how can we convert the ``data work'' which is often undervalued as merely operational and distil these ad hoc processes into systematic data-centric tools? In addressing the highlighted issues and curating high-quality datasets, the goal is to empower rather than replace data scientists/ML engineers by making processes more systematic, making processes less prone to human variability, and freeing up human experts for other tasks.

\paragraph{\bf Q1: How did you select, collect or curate your dataset?}
ML datasets are often curated or selected in a one-off  ``benchmarking'' style fashion. We aim to raise awareness, that without careful consideration, this approach has the following potential pitfalls; (1) fixation on performance on these datasets which might not translate to real-world performance \cite{tsipras2020imagenet}, (2) these datasets are imperfect and have labeling errors / inconsistencies and biases \cite{northcutt2021pervasive, gebru2021datasheets}, which can influence the learning algorithm (especially true if labels are generated via crowdsourcing) and (3) once-off collection does not represent the real world. These three issues mean that, while model performance on these datasets is laudable, it does not provide real-world guarantees of performance, even for the same task. Cases such as degradation from ImageNet V1 to V2 highlight this issue, even in controlled settings \cite{shankar2020evaluating}. 

We do not suggest that curated datasets should be completely discarded. On the contrary, they have contributed greatly towards model advancements and are a good starting point for many applications. Instead, when constructing and curating datasets, these pitfalls should be taken into consideration and addressed.


A further consideration is that high-quality datasets in academia or industry are often collected ``one-off'', requiring huge amounts of human labeling effort. However, in real-world settings, models need to be frequently updated with new data (potentially daily). Even if the problem might seem static (e.g. detecting pneumonia in an x-ray), there is still a need for data updates. This raises the challenge of how to systematically curate new datasets over time, without the need for expensive human labeling. We discuss potential data-centric opportunities later to address this.

Beyond simply curating a high quality, consistent labeled dataset; we draw attention to the need for forensics on dataset selection and what qualities make a dataset useful for a specific task. Specifically, this concerns whether the data is indeed pertinent and  sufficiently representative. In particular, we want to avoid situations similar to the early stages of the Covid-19 pandemic, where researchers unknowingly used a common chest scan dataset from \cite{kermany2018identifying} as a control group against actual Covid-19 positive scans. However, as many did not realize, the control dataset consisted of paediatric patients. Hence, while there was an illusion of reliable performance in detecting Covid-19, the reality is that the data mismatch transformed the model into an adult vs child detector \cite{roberts2021common}. While seemingly an extreme case, we believe that similar scenarios are entirely avoidable with a data-centric lens from the offset, wherein datasets are audited for such subgroup dependencies and interactions. Although data assessment to prevent this would be largely manual right now, we present potential opportunities to improve this process.

Related to forensics are considerations of bias introduced by humans through the data collection process, but also how systems themselves can bias data through degenerate feedback loops \cite{jiang2019degenerate}. For example, a system's output, such as a recommendation system, can influence the user's behavior. The new behavior can then contribute to ``degeneracy'' in the new training dataset. Not accounting for this subtlety in how biases can be injected and insufficiently auditing them from the offset can lead to unintended behaviors. We believe that concepts of data provenance could play a role in this regard. Next, we highlight opportunities to address these challenges.

\textbf{\textit{Opportunities:}}
First, \emph{continuous dataset curation} without large amount of human labeling. Advances in weak supervision \cite{ratner2016data} or even semi-supervised labeling functions based on automatic labeling, coupled with rule-based systems as has been used in NLP could allow this realization \cite{smit2020combining}. Second, \emph{merging data sources}, from multiple different sources, e.g. multi-modal ML (e.g. fusing images with text etc) or fusing disparate datasets with similar signal. For example, datasets from different countries or datasets with different features for the same modality. Third, \emph{automated data forensics}, to prevent unintended consequences for ML models. This could involve identifying inappropriate subsets of the data or degenerate data. While this is a currently manual process, there exists opportunities if not to automate this, but to develop human-in-the-loop methods to identify candidate problematic data.

\paragraph{\bf Q2: What data cleaning and/or pre-processing, if any, has been performed?}
Typical steps of the colloquial term ``data wrangling'' and cleaning \cite{endel2015data} could involve one or more of the following (for brevity we assume the data engineering pipeline and quality control has been accounted for): (1) Data cleaning and pre-processing: obvious errors such as outliers, mislabelled or implausible data due to system level errors \cite{chu2016data,tae2019data} (i.e. data engineering), (2) Feature selection \cite{jovic2015review,li2017feature} and engineering, either via intuition or domain knowledge (such as including key clinical predictors or business features) and (3) Data imputation of missing data \cite{efron1994missing, jadhav2019comparison}.  Failure to consider such data issues can at best lead to a suboptimal model and at worst capture the incorrect statistical relationship (e.g. \cite{krishnan2016activeclean}).

\textbf{\textit{Opportunities:}} The aforementioned processes are largely manual and rely on human intuition. Opportunities exist to build data-centric tools to make this more systematic and autonomous. AutoML for data cleaning or reinforcement learning agents which select the components of such a processing pipelines are potential avenues ripe for exploration.

\paragraph{\bf Q3: Has data quality been assessed?}
The quality of the dataset affects our ability to learn. How can we quantify signal at a dataset level, especially in determining upper-bounds on expected performance? Specifically, how to identify high-quality samples that are ``easy'' to learn from and differentiate from more challenging or lower-quality samples where additional features might be required, the statistical relationships may not hold, or the sample might be mislabelled. Applicable methods include those that assess intrinsic \emph{instance hardness} \cite{smith2014instance}, analyzing how individual examples evolve differently based on model \emph{training dynamics} \cite{swayamdipta2020dataset,pleiss2020identifying} or explicitly modelling the \emph{data value} of examples \cite{yoon2020dvrl}.

\textbf{\textit{Opportunities:}} Current approaches to assessing data quality can characterize samples in a dataset as easy, ambiguous, or hard. However, most methods are model-dependent (even for models that perform similarly). Consequently, the model itself might influence the characterization of individual data samples. Ideally, we desire a data-centric approach where the findings are inherent to the difficulty of the data itself, rather than the model used to assess the data (of course, assuming the models are similarly performing). The importance is to ensure that the instances identified represent the inherent difficulty with the data (aleatoric) rather than the model's challenge with the data (epistemic). The utility of this task, beyond assessing the quality of data samples, is to either sculpt the dataset by filtering the challenging data points or rigorously test performance on these challenging data points/subgroups.

\paragraph{\bf Q4: Have you considered synthetic data?}
Synthetic data generation \cite{nikolenko2021synthetic} typically involves generative models to synthesize data, often considered for privacy-preserving reasons or due to data access limitations \cite{emam2020}, especially in regulated settings (e.g. healthcare or finance). Thus, synthetic data might be considered if such considerations are relevant.  For situations where this is not the case, we ask if synthetic data could provide utility – for this, we detail opportunities. We note the difference between synthetic data generation and data augmentation. Synthetic data refers to new data points that did not exist in the dataset, whilst data augmentation is a new version of an existing data point.

\textbf{\textit{Opportunities:}} Opportunities for synthetic data lie beyond access and privacy issues. e.g. improving data quality by increasing dataset coverage. Alternatively, generating fair and unbiased datasets reflecting the desired data generating process, which might differ from reality. Finally, generating data of specific scenarios we want to stress test.

\section{\textbf{Training}: a data-centric lens informing model training}
\begin{figure}[!h]
    \centering
    \includegraphics[width=0.9\columnwidth]{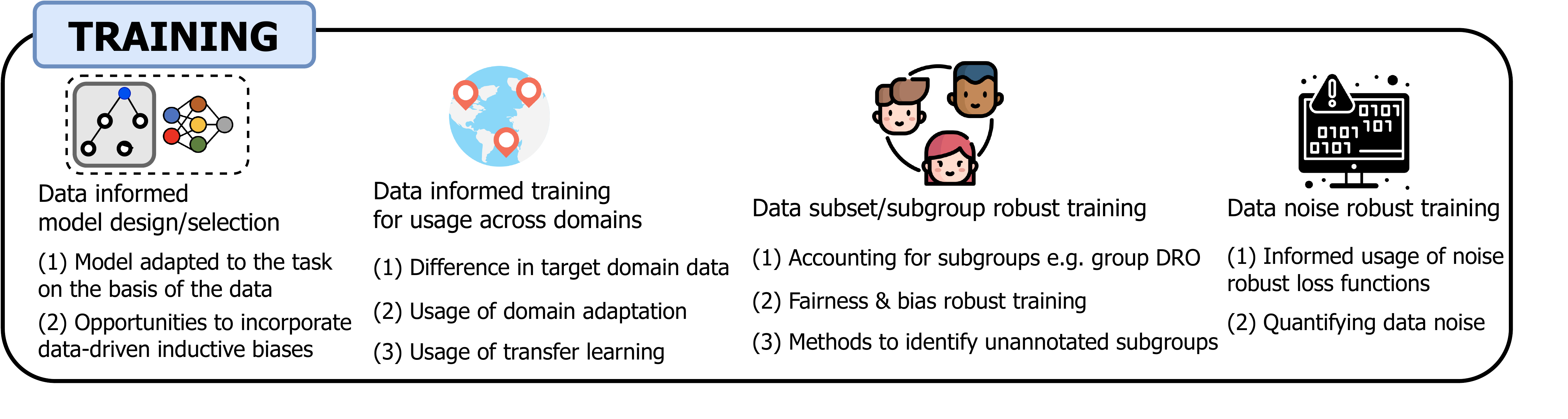}
    \caption{TRAINING: considerations for more systematic data-centric informed model training}
    \label{fig:training}
\end{figure}
How can we ensure reliable model training? While this might sound inherently model-centric as we modify the model training, we motivate the role of data in informing model training. For instance, by understanding the data, one might highlight the need for robust training procedures. In addition, specific remedies often require knowledge of the desired properties and data distributions. Clearly, this is different from conventional model-centric notions of improving the architectures, changing model layers, or tuning hyperparameters. Although, these remain important concepts.

\paragraph{\bf Q5: Have you conducted a model architecture and hyperparameter search?}
There is ``no free lunch'' when it comes to selecting a ML model for a particular task. Thus, it is important to compare a variety of models, including ``simple'' alternatives. We could also consider how the data could inform the architecture choice. For example, attention layers work well in NLP due to the nature of text data. Going beyond model search, documenting hyperparameter search is also crucial for reliable ML pipelines.

\textbf{\textit{Opportunities:}}  Understanding the data could better inform the model architecture itself, allowing one to incorporate inductive biases \cite{baxter2000model} using causal models, domain knowledge, and expert models. Such understanding could also guide the hyperparameter search to make it less exhaustive. e.g. hyperparameters translating across similar datasets.

\paragraph{\bf Q6: Does the training data match the anticipated use?}

Thinking critically about the possible environments in which the model will be deployed, and whether the training data is sufficient, is important. Consider the case where the domain or distribution differs from the training data.  Failure to consider such data issues can lead to trained models with poor generalization capability or providing biased estimates. To address this in cases where it is not possible to curate or select more appropriate training data, one could consider transfer learning \cite{kouw2018introduction,zhuang2020comprehensive} and domain adaptation \cite{kouw2018introduction,ganin2015unsupervised,zhao2019learning}, where the goal is to adapt the model or representation space to match the desired use. 

\textbf{\textit{Opportunities:}} The aforementioned methods are often predicated on the assumption that data from the target domain is readily available. Hence, opportunities exist to develop methods to reduce the reliance on data from a target domain - i.e. domain adaptation or transfer learning with limited data from the target domain. This could be useful, for example, in low-resource healthcare settings. Another area to consider is transfer learning. Although these methods are often used for ``unstructured'' data such as images and text, there exist opportunities to formalize how we can transfer in tabular settings; in particular, in cases with feature mismatch.

\paragraph{\bf Q7: Are there different data subsets or groups of interest?}

There are often specific groups of interest (e.g. high-risk patients). Alternatively, we might want to ensure fairness and/or robustness across different subsets of data. Such considerations are often important in high-stakes decision making settings, such as healthcare or finance.  With such considerations in mind, one could consider employing training methods with fairness objectives \cite{corbett2018measure,friedler2019comparative} or methods that directly optimize to achieve group-wise performance parity \cite{rahimian2019distributionally,sagawa2019distributionally}, i.e., instead of empirical risk minimization to maximize average performance, the goal is equalizing performance across groups, or minimizing the worst group error. 

\textbf{\textit{Opportunities:}}  
Current methods for optimizing group-wise parity typically need group labels, which are not always available in the data. Hence, there are opportunities to develop data-centric methods to identify such regions or subsets of data when group labels are unavailable. Additionally, these methods typically come at the cost of sacrificing overall accuracy. Novel methods should be explored to mitigate the loss of overall performance. For example, as outlined by \cite{roh2021sample}, addressing fairness in subsets could harm accuracy and vice versa; improving accuracy could reduce fairness. Novel methods to better balance trade-offs are of great need.

\paragraph{\bf Q8:  Is the data noisy, either in features or labels?}

Collected data points might be noisy and/or mislabelled. This is especially true where data labels are obtained through crowd-sourcing or automated methods. Identifying the presence of such artifacts can then motivate the use of methods to learn with noisy data. These could include noise robust loss functions, including architecture layers robust to noise, meta-learning, or dataset sample selection \cite{song2020learning}. 

\textbf{\textit{Opportunities:}}  A myriad of methods aim to solve the noisy data learning problem (see \cite{song2020learning}) and, of course, can be further improved. In addition, a significant opportunity lies in the data-centric task of identifying and quantifying that indeed the dataset consists of noisy or mislabelled samples, and hence the aforementioned methods are required.

\section{\textbf{Testing}: a data-centric lens informing new approaches to testing}

\begin{figure}[!h]
    \centering
    \includegraphics[width=0.85\columnwidth]{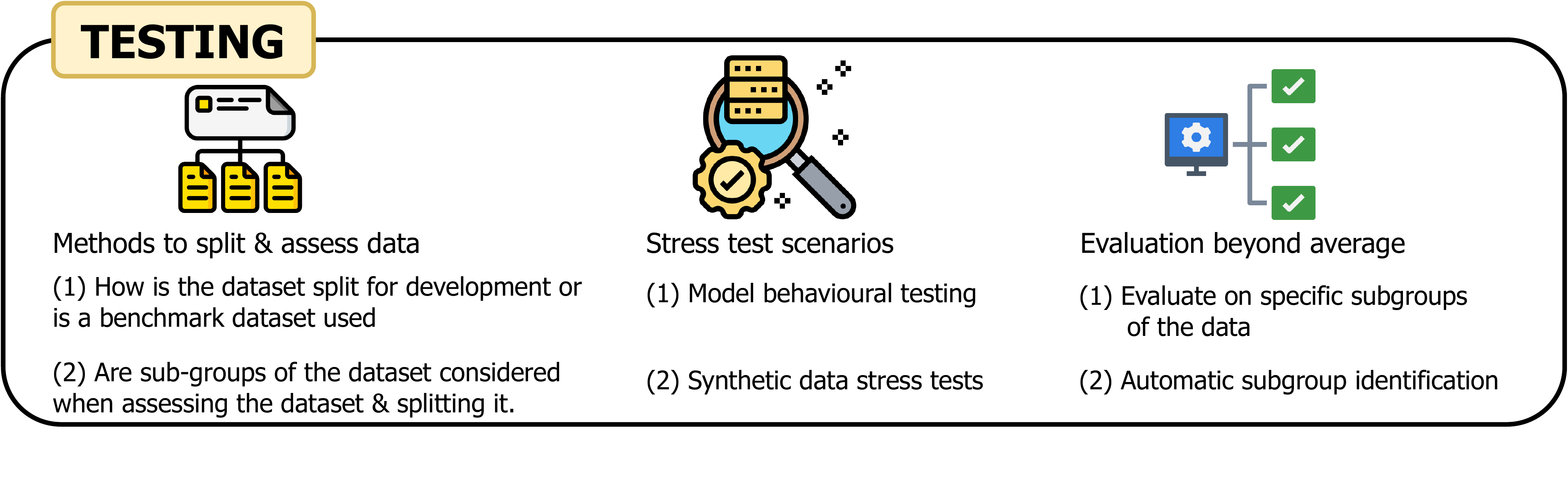}
    \caption{TESTING: considerations where a data-centric lens informs  testing }
    \label{fig:testing}
\end{figure}

Before deploying trained ML models, they need to be rigorously tested. Mature industries involve standard methods to benchmark and provide characterizations of components under a variety of testing and/or operating conditions \cite{gebru2021datasheets}. For instance, automobiles make use of wind tunnels and crash tests to assess specific components, while electronic component datasheets outline conditions where reliable operation is guaranteed. On the contrary, current approaches to characterize performance of ML models do not have the same level of detail and rigor.  The current de facto in ML is benchmarking style, where the model is evaluated on a held-out test set and assessed on average/aggregate performance. Can we do better and provide levels of rigor and detail for ML testing? We outline the considerations below.

\paragraph{\bf Q9: How has the dataset been split for model training and validation?}
Thoughtful and careful documentation of data splits for training and validation are often overlooked. e.g. random splits might not always be optimal, especially if the curated datasets might themselves not be IID. This can happen when datasets are collected over time with seasonality or evolving patterns. Blindly aggregating the data and ignoring the temporal nature when splitting the data can at best mask phenomena and at worst mean that we might capture idiosyncrasies over time. Addressing this issue is obvious when temporal data is available. However, certain datasets such as image datasets often do not contain this information, which might mask such phenomena. Another issue surrounding data splitting is whether subgroups are actually considered. Random splits might inadvertently introduce selection biases or coverage gaps. Assessing representativeness in data selection and splitting is vital to mitigate such issues. 
Often careful consideration should be given to the split used depending on both the specific problem and the use-case for the model. For example, in drug discovery, scaffold splits, where structurally different molecules are kept separate might be necessary for one use-case, but less appropriate for another \cite{Wu2018}; similarly, cross-target splitting is necessary for models intended to be deployed on novel proteins, but less important for other applications \cite{imrie2018protein}.

\textbf{\textit{Opportunities:}} Automatic methods that characterize/identify and subsequently account for subgroups are an avenue for future work. Especially in terms of assessing the representativeness of data splits. Of course, methods such as cross-validation might mitigate such issues in small cases, but new methods are needed to scale to large datasets.

\paragraph{\bf Q10: How has the model been evaluated (e.g. metrics \& stress tests)?}

Related to the issue around using benchmark datasets, we stress the importance of both how and on what data the model will be evaluated. Typical approaches might use predefined train-test splits (benchmark) or randomly split the dataset. However, we advocate for additional \textit{scenario-based} evaluation.  i.e. test scenarios, motivated by real-world considerations that probe specific aspects of model performance.  For example: face recognition could assess performance for different scenarios such as demographic, lighting conditions etc. Identify model failure scenarios, could guide where additional targeted data collection might help to improve the model. Additionally, ML models are typically evaluated on average on a testing dataset. We highlight the importance of assessing model performance on subgroups of the data, to ensure that indeed a model is performant on all groups of the data. In particular, there are no hidden subgroups or hidden stratification \cite{oakden2020hidden} of the data in which a model may systematically under-perform. This relates to Q7 about subsets or groups of interest during training. One approach is subpopulation testing such as race, gender age etc, long tail events or temporal distribution shift.  e.g. \cite{dulhanty2019auditing} shows performance differences across races for computer vision models on the PPB dataset.

Finally, thoughtful definition of metrics is critical to useful evaluation. e.g. in imbalanced data scenarios, one would prefer to assess the model using precision or recall instead of accuracy, which might overestimate the model's true predictive abilities. Documenting the choice of evaluation metric and the rationale is a step in the right direction \cite{mitchell2019model}.

\textbf{\textit{Opportunities:}} While benchmark datasets and competing based on performance have their place, can we better evaluate models at a granular level?  Naturally, we could either partition an existing dataset that might suffer from small sample sizes or collect data across different scenarios, which is labor intensive. Consequently, we draw attention to the role that synthetic data could play to generate data to stress test specific scenarios, without having to collect data from said scenarios. e.g. testing the model's robustness under covariate shift. In terms of subgroup characterization, current approaches rely on the assessment of specific, well-known subgroups in the data. However, what if a subgroup is not well-defined or is intersectional, yet we still wish to identify the underperforming slice. There are obvious subgroups we can assess (e.g. race, sex etc). However, a research challenge is to identify subgroups that perform poorly and are not tied to well-known features or might have intersectional factors.
Hence, opportunities exist to develop data-centric methods for robust and automatic subgroup characterization (slice-discovery) as a way to systematically assess model performance on different regions of the data. Opportunities always exist for new metrics, since there isn't one metric to rule them all. From a data-centric lens, this could be calibrating metrics to the difficulty of the dataset, or the subset evaluated over.  Additionally, work is needed on better proxies for what humans are measuring (e.g. business KPI).

\section{\textbf{Deployment}: a data-centric lens informing reliable and observable deployment }

\begin{figure}[h]
    \centering
    \includegraphics[width=0.9\columnwidth]{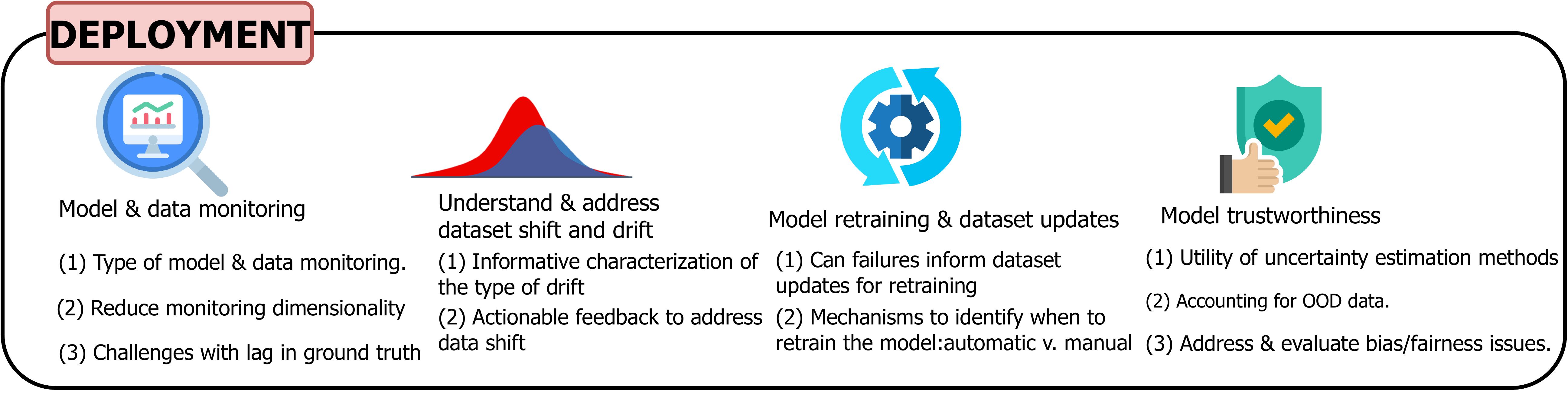}
    \caption{DEPLOYMENT: data-centric considerations  for deployed ML models}
    \label{fig:deployment}
\end{figure}

We can only realize the value of predictive models once they are deployed into the wild. This then evokes additional considerations to ensure reliability. In this section, we do not consider systems and infrastructure challenges, typically covered under the umbrella of MLOps, along with the vast body of software engineering and DevOps work. Instead, we focus on ML-specific considerations. First, we wish that the models provide reliable predictions when operated under a variety of conditions. Additionally, we expect that models ``know what they do not know'', especially when dealing with uncertain or out-of-distribution data.  Second, the world is not stationary, and hence models will naturally degrade over time. Thus, one might want to include monitoring and observability to data and distribution shifts as a means to ensure system reliability. 
A data-centric lens at deployment time consists of characterizing incoming data, in order to assess if the model has the desired capabilities to make reliably predictions on said data. This brings up the conundrum, in the case where the data has changed, of both adapting the  data and subsequently re-training the model. This step closes the feedback loop between DATA and TRAINING in a data-informed manner.

\paragraph{\bf Q11: Are you monitoring your model?}
While seemingly obvious, establishing a systematic process to monitor a deployed model is a crucial but often overlooked step. 
Monitoring can manifest in two main ways: (1) detecting divergences in the data and (2) mechanisms to detect the failures of the model (ideally with statistically guarantees). 
Monitoring the deployment distribution is crucial, as training and deployment distribution mismatch is a significant cause of ML system failure. The result is models behaving unpredictably or fail silently and issuing incorrect, yet overconfident predictions, making them challenging to audit. Hence, , monitoring and observability solutions could consider methods to detect dataset shift and methods to address the shift.  However, before examining each, we define the three most prevalent shifts \cite{quinonero2008dataset}. Further, while shifts might seem like step-like, instantaneous events, the rate of shift can vary in reality. This variability impacts our ability to detect it, with gradual shifts more difficult to characterize.

\begin{itemize}
 \item \textbf{Covariate shift}: $P(X)$ changes, $P(Y|X)$ remains the same. e.g.  diabetic retinopathy screening: training data has high quality images. However, when deployed there is a covariate shift with images captured of lower quality.  
 \item \textbf{Label shift}: $P(Y)$ changes (i.e. prior shift), $P(X|Y)$ remains the same. e.g. we train an animal image classifier on North American animals. When used in Africa the distribution of target animals would have shifted.
 \item \textbf{Concept drift}: $P(Y|X)$ changes, $P(X)$ remains the same. e.g. predicting customer spend. Covid-19 then changes customer behavior, affecting the statistical properties of the input-output relationship (i.e. posterior shift).
\end{itemize}

The most basic monitoring procedure is to monitor the model's performance metrics, compared to expected performance or a baseline, i.e. service level objective, such as accuracy. However, in many real-world scenarios, ground truth labels are often either unavailable or have time lags in availability, which necessitate alternative monitoring approaches. 

One solution is to monitor the data pipeline by data schema validation or placing constraints on the data values \cite{breck2019data}. Another approach is data drift detection, which monitors the input or predicted label distributions and assesses if there is a discrepancy from training data. One could consider simple statistics approaches to more advanced distributional divergence methods. 
The aforementioned approaches assume a single train and single test set comparison. i.e. this simplifying assumption distills distribution shift as a comparison of two finite sets of data. In reality, an ML model is deployed and makes predictions over time. Hence, we need to assess drift in a streaming setting with respect to a specific time horizon. Thus, selecting the time horizon width to segment and compute potential drift is non-trivial. One could consider whether the perceived drift is likely to be gradual or rapid in  determining an optimal slicing interval.

\textbf{\textit{Opportunities:}} Approximating model performance without labels (e.g. due to lag), is an impactful opportunity. If this is not possible, we still want to detect drift. Current methods are largely suited to low-dimensional data, and hence new methods tailored to high-dimensional settings (both tabular and non-tabular) are needed. Finally, principled approaches to optimal slicing intervals would greatly improve streaming detection.


\paragraph{\bf Q12: Do you have mechanisms in place to address data shifts?}

Once data shift has been detected, one could consider methods to address the shift, such as adapting the model.  A naive approach might attempt to train the model on a sufficiently large dataset in the hope of covering the entire distribution which we might encounter in the wild. This has easy points of failure if the full distribution is not truly captured, and is only applicable in limited scenarios where large datasets are readily available. The second approach previously discussed under training is learning domain-invariant representations. This approach is one example of a data-centric lens informing model-centric development. Finally, we could adapt or retrain the model using data from the new distribution. However, there are a number of factors to consider: how can we efficiently construct a new labeled dataset,  when is the optimal time to retrain,  how often should a model be updated (i.e. cadence) and whether to use a batch-training approach vs continual learning approach. Answering these questions efficiently is covered in the opportunities section.

\textbf{\textit{Opportunities:}}  
An overlooked opportunity is how to construct new datasets for retraining. Currently, new datasets are constructed manually and are also done so without failures informing the dataset updates. We propose feedback-driven datasets, with a data-centric enabled feedback loop. By this we mean errors informing the automatic creation, updating, or augmentation of new training sets. In fact, the type of "self-tuning" could be informed in a data-centric manner by the type of drift. e.g. concept drift requires retraining on the latest data, whilst covariate shift we could augment/synthetically generate certain subsets. Additionally, methods to automatically determine the optimal retraining intervals would be hugely impactful in industrial settings where models are updated on a daily/weekly cadence.

Furthermore, what happens when many features are flagged as having drifted? It is often practically unactionable in such cases. For example, if tens or hundreds of features have high KL divergence values. Hence, new methods to actionably identify the root cause of drift are opportunities for development. Furthermore, we wish that the feedback is human understandable with recourse beyond simply highlighting features that have changed. 

\paragraph{\bf Q13: Have you incorporated tools to engender model trust?}
For models to produce trustworthy predictions, we desire that models reflect their uncertainty when predicting on certain data, as well as identify when they are required to predict on data outside its realm of expertise. One could consider the myriad of uncertainty quantification techniques (e.g. Bayesian neural networks, conformal prediction, Gaussian processes, MC Dropout etc.) to ascertain a model's predictive uncertainty. Uncertainty quantification could then be used to defer predictions if they are too uncertain – ensuring reliability by selective prediction.
Uncertainty can also provide actionable feedback whether to trust predictions. Beyond uncertainty quantification; methods from explainability (e.g. feature importance, concept-based etc) and algorithmic fairness (e.g. Equalized odds postprocessing) could arguably also fall into this category to engender trust in a deployed model. Finally, models should identify when asked to predict on data outside its training distribution. Out-of distribution (OOD) detection can play a role in this case, (e.g. likelihood ratios, distance based or density based). However, even if a data point lies within the support of the distribution, it could lie in a region where a model might make poor predictions. Recent work such as \cite{seedat2022data} could identify such inconsistent and incongruous examples of in-distribution data.

\textbf{\textit{Opportunities:}} There are opportunities for improved methods of uncertainty quantification, model explainability, algorithmic fairness and OOD detection. Additionally, not all errors are created equally, with some errors more harmful than others. Ideally, we want to assess if a model should be trusted uniformly or whether specific data subgroups have greater uncertainty, have different model explanations, are unfairly prejudiced or contain more OOD samples. 

\begin{table*}[!h]
\centering
\caption{Summary of current approaches with opportunities outlined in DC-Check. Green represents ideas unique to DC-Check.}
\vspace{0.5cm}
\scalebox{0.99}{
\begin{tabular}{|c|c|}
\toprule
Current & DC-Check                       \\ \midrule
 \multicolumn{2}{|c|}{DATA}\\\midrule
Benchmark/Highly curated datasets &  Proactive selection/curation \\
Fixed datasets & \cellcolor{green!10} Continuous dataset curation \\
Manual data forensics &  \cellcolor{green!10} Automated data forensics  \\ 
Ad hoc data pre-processing & \cellcolor{green!10}  Systematic data cleaning tools (AutoML/RL agents) \\
Manual dataset improvement & \cellcolor{green!10}  Synthetic data beyond privacy preservation \\ \midrule
 \multicolumn{2}{|c|}{TRAINING}\\\midrule
Performance based model architecture search & \cellcolor{green!10} Data informed architecture selection \\
Heuristic/manual robust learning & \cellcolor{green!10} Data informed robust learning \\ 
Domain adaptation and transfer learning & Improving these methods for limited data \\
Fairness and group robust methods & Methods to balance fairness/robustness with performance \\
Learning robust to noisy data & Data-centric informed usage of such methods \\
\midrule
 \multicolumn{2}{|c|}{DEPLOYMENT}\\\midrule
Limited or Low-dimensional monitoring & \cellcolor{green!10}  New methods for high-dimensional moonitoring \\
Naive data shift remedies  & \cellcolor{green!10} Actionable and understandble shift remedies \\
Naive model retraining (batch)  & Continual learning (streaming) \\
Naive dataset updates  & \cellcolor{green!10} Selt-tuning datasets \\
Overconfident models & Uncertainty estimation \& OOD detection   \\ \midrule
 \multicolumn{2}{|c|}{TESTING}\\\midrule
Fixed data evaluation & \cellcolor{green!10}  Synthetic stress test based evaluation  \\
Average/population-level evaluation & \cellcolor{green!10}  Subset/subgroup evalution evaluation  \\\midrule
\bottomrule
\end{tabular}}
\label{related_work}
\end{table*}

\section{Going beyond supervised learning with DC-Check}

The DC-Check framework is intended to be usable across a variety of contexts, use-cases and stakeholders. Thus, DC-Check has been framed in the context of supervised learning, it is of course not the only applicable setting.  We believe that the scope of issues and opportunities are flexible enough and arise across a wide variety of ML use-cases, from unsupervised learning to areas such as causal inference. Of course, the DATA area is the most natural fit when moving beyond supervised learning. In particular, systematic methods to curate and assess datasets are applicable across all strands of ML. TRAINING naturally is field specific. However, for DEPLOYMENT and TESTING much more work is required to develop better methods and practices to translate beyond supervised learning. Overall, we wish to highlight that the broad data-centric considerations and opportunities posed in DC-Check are nevertheless important for both practitioners and researchers alike in developing reliable end-to-end ML pipelines, no matter the ML paradigm.

\newpage
\section{Bridging the reliability gap with a data-centric lens}
We present \textbf{DC-Check}, a checklist framework with a data-centric lens, as a guide toward reliable ML systems at a pipeline level.   DC-Check is intended for both ML researchers and practitioners to leverage in day-to-day development. We believe that the \emph{data-centric} spirit is key to take us from the ad hoc world where reliability considerations are an afterthought prior to deployment (or shockingly after deployment failures), and transport us across the reliability divide to a world where these considerations are systematic and baked into the development process from the get-go.

The time is indeed highly pertinent for the ML community to reflect on considerations of reliable ML systems. We have passed "the making ML model work phase" to the "making real-world ML systems phase". No longer can we simply optimize for predictive performance on benchmark tasks and assume success in the real-world.  This is evident as ML applications have become widespread across industries. Furthermore, the fact that high-stakes settings of healthcare and finance are beginning to implement such tools, reliability considerations should clearly be standard rather than after thought. Without considering the many other interconnected factors outlined by frameworks such as DC-Check, we run the risk of potentially impactful failures in the real world. 

We hope that DC-Check's actionable steps, coupled with the broader interaction of data-centric AI, will spur the research community to address the opportunities to realize the goal of highly performant and reliable ML systems.

\bibliographystyle{unsrt}  
\bibliography{references}

\clearpage
\appendix

\section{Interconnected components of DC-Check}
\begin{figure}[!h]
    \centering
    \includegraphics[width=0.9\columnwidth]{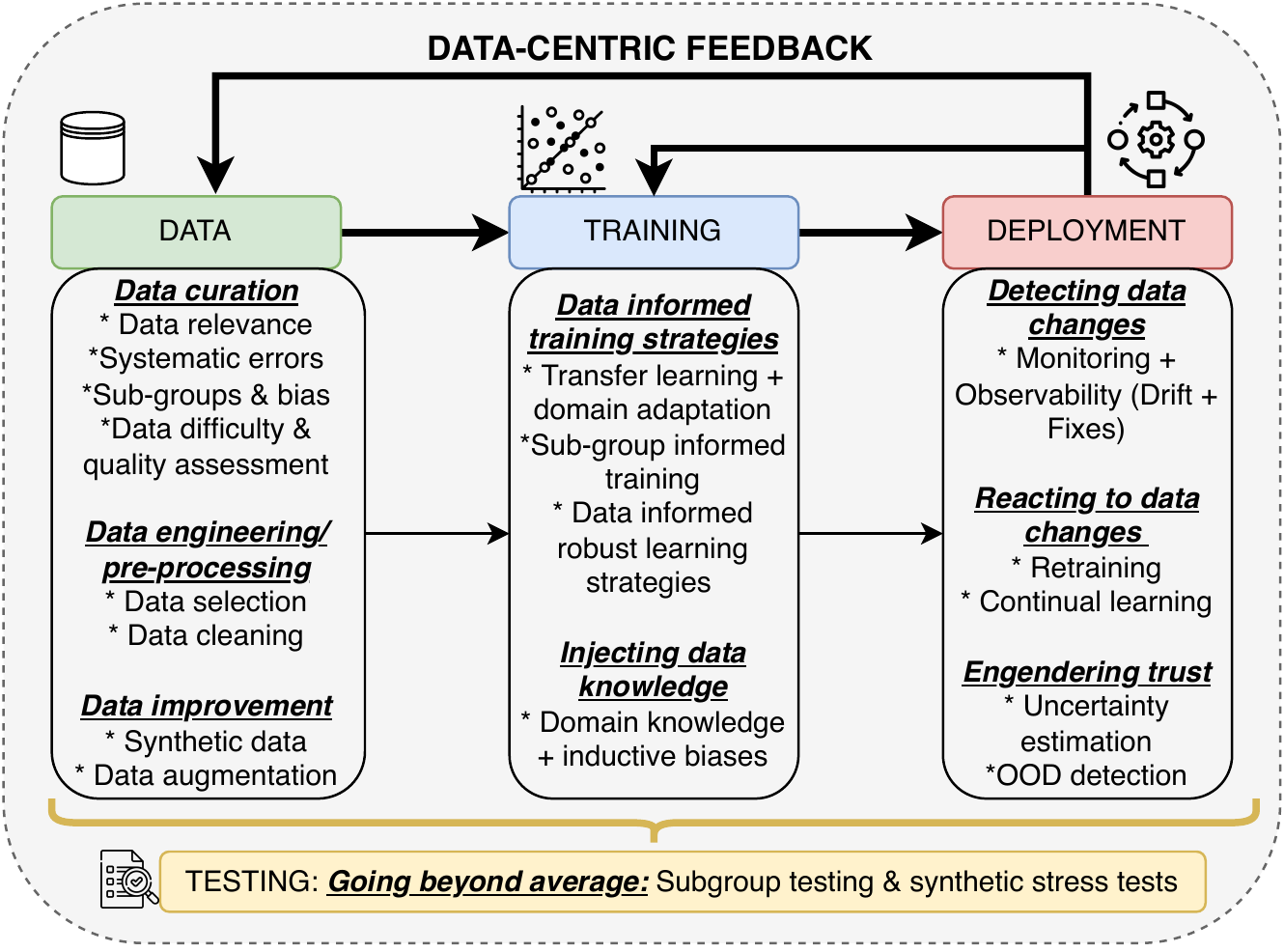}
    \caption{Overview of the interconnected flow of \emph{DC-Check} across the pipeline}
    \label{fig:overview}
\end{figure}

\newpage
\section{Example tools for DC-Check}

This appendix provides a set of example tools, are by no means exhaustive, that could be used to address in each section of DC-Check. DATA is in Table \ref{tools-data}, TRAINING is in Table \ref{tools-training}, TESTING is in Table \ref{tools-testing}, DEPLOYMENT is in Table \ref{tools-deployment}.
\begin{table*}[!h]
\centering
\caption{Example Tools at the Data Stage.}
\scalebox{0.95}{
\begin{tabular}{|c|c|}
\toprule
Task & Example Tools                       \\ \midrule\midrule
\multicolumn{2}{|c|}{ \cellcolor{green!10} DATA}\\\midrule\midrule
 
\multicolumn{2}{|c|}{\bf Q1: How did you select, collect or curate your dataset?}\\\midrule

Data curation  &  \makecell{N.A. - Human-driven} \\ \midrule

Data selection  &  \makecell{N.A. - Human-driven} \\ \midrule

Data forensics  &  \makecell{N.A. - Human-driven} \\ \midrule
  
\multicolumn{2}{|c|}{\bf Q2: What data cleaning and/or pre-processing, if any, has been performed?}\\\midrule

Data cleaning \& pre-processing  & \makecell{Outlier removal \\ Schema Validation (Great Expectations) \\ Constraints violations \\ Remove duplicates \\Correct systematic errors \\ Syntax errors \\ Normalization/Standardization \\ Type conversions}\\ \midrule

Feature selection  &  \makecell{Filter-based \\
Wrapper-based \\
Embedded} \\ \midrule

Data imputation   &  \makecell{Mean imputation, KNN-based \\ MICE \citeA{royston2011multiple} \\MissForest \citeA{stekhoven2012missforest} \\ GAIN \citeA{yoon2018gain}} \\ \midrule
    
\multicolumn{2}{|c|}{\bf Q3: Has data quality been assessed?}\\\midrule

Quality/type of data samples  &  \makecell{AUM \citeA{pleiss2020identifying} \\ Data Maps \citeA{swayamdipta2020dataset} \\ Instance Hardness \citeA{smith2014instance} \\ DVRL \citeA{yoon2020dvrl} \\ Data Shapley \citeA{ghorbani2019data}}\\ \midrule

Labeling error  &  \makecell{AUM \citeA{pleiss2020identifying} \\ CleanLab \citeA{northcutt2021pervasive}} \\ \midrule
        
\multicolumn{2}{|c|}{\bf  Q4: Have you considered synthetic data?}\\\midrule

Synthetic data for privacy/access  &  \makecell{PateGAN \citeA{jordon2018pate}, ADSGAN \citeA{yoon2020anonymization}, DP-cGAN \citeA{torkzadehmahani2019dp}}\\ \midrule

Synthetic data for augmentation  &  \makecell{CTGAN \citeA{xu2019modeling}\\ HealthGen \citeA{bing2022conditional} \\ DAGAN \citeA{antoniou2017data}} \\ \midrule

Synthetic data for quality  &  \makecell{DECAF \citeA{van2021decaf}} \\

\bottomrule
\end{tabular}}
\label{tools-data}
\end{table*}

\begin{table*}[!h]
\centering
\caption{Example Tools at the Training Stage.}
\scalebox{0.95}{
\begin{tabular}{|c|c|}
\toprule
Task & Example Tools                       \\ \midrule\midrule

 \multicolumn{2}{|c|}{ \cellcolor{red!10} TRAINING}\\\midrule\midrule
 
 \multicolumn{2}{|c|}{\bf Q5: Have you conducted a model architecture and hyperparameter search?}\\\midrule
 
 Architectures  &  \makecell{Pre-trained architectures \citeA{he2016deep,devlin2018bert}\\ Attention layers \citeA{vaswani2017attention}} \\ \midrule
 Hyper-parameters &  \makecell{Best practices \& Search types \citeA{bergstra2013hyperopt, hertel2018sherpa}} \\ \midrule
 
 \multicolumn{2}{|c|}{\bf Q6: Does the training data match the anticipated use?
}\\\midrule

Transfer learning  &  \makecell{Fine-tuning, CORAL \citeA{sun2017correlation}, DAN \citeA{long2015learning}, DANN \citeA{ganin2016domain}} \\ \midrule

Domain adaptation  &  \makecell{Domain-invariant layers \citeA{zhao2019learning} \\ Domain adversarial training \citeA{ganin2016domain}} \\ \midrule
  
  \multicolumn{2}{|c|}{ \bf Q7: Are there different data subsets or groups of interest?}\\\midrule
  
  Subgroup robust training  &  \makecell{Group-DRO \citeA{sagawa2019distributionally}  \\ Group weighted losses}\\ \midrule
   
 \multicolumn{2}{|c|}{\bf Q8:  Is the data noisy, either in features or labels?
}\\\midrule

Robust losses  &  \makecell{Active Passive Loss \citeA{ma2020normalized}\\ Bi-tempered Loss \citeA{amid2019robust} \\ Generalized Cross Entropy \citeA{zhang2018generalized}} \\ \midrule

Robust layers  &  \makecell{Noisy Adaptation
Layers \citeA{sukhbaatar2014training} \\ Noise networks \citeA{patrini2016loss}} \\ \midrule

Sample selection  &  \makecell{Mentornet \citeA{jiang2018mentornet},CoTeaching \citeA{han2018co}, Iterative Detection \citeA{wang2018iterative} } \\

\bottomrule
\end{tabular}}
\label{tools-training}
\end{table*}

\begin{table*}[!h]
\centering
\caption{Example Tools at the Testing Stage.}
\scalebox{0.95}{
\begin{tabular}{|c|c|}
\toprule
Task & Example Tools                       \\ \midrule\midrule

 \multicolumn{2}{|c|}{ \cellcolor{blue!10} TESTING}\\\midrule\midrule
 
 \multicolumn{2}{|c|}{\bf Q9: How has the dataset been split for model training and validation?}\\\midrule
 
 Data splits  &  \makecell{Random splits \citeA{dangeti2017statistics}, Cross-validation\citeA{dangeti2017statistics},\\ Temporal splits \citeA{cerqueira2020evaluating} ,Task specific splits \citeA{hu2020open} }\\ \midrule
 
 \multicolumn{2}{|c|}{\bf Q10: How has the model been evaluated (e.g. metrics \& stress tests)?
}\\\midrule\midrule

Metrics  & \makecell{Task dependent metrics\citeA{hossin2015review}\\ Imbalance robust metrics (AUC, F1-Score etc) \citeA{hossin2015review,fatourechi2008comparison} \\ Per-group metrics \citeA{sohoni2020no} \\ Worst-group metrics \citeA{sohoni2020no}}\\ \midrule

Stress testing  &  \makecell{Testing under different conditions (e.g. lighting, environments) \citeA{buolamwini2018gender,hendrycks2018benchmarking} \\ Testing under different model inputs \citeA{ribeiro2020beyond} \\ Testing with data containing spurious artifacts \citeA{mahmood2021detecting,veitch2021counterfactual}} \\ \midrule

Subgroup testing  &  \makecell{Testing specific subgroups/subpopulations \citeA{buolamwini2018gender}  \\ Cluster-based or discovered subgroups \citeA{herrera2011overview,zimmermann2009cluster}\\ Use-case specific subgroups \citeA{irvin2020incorporating,ye2018prediction}} \\ 
\bottomrule
\end{tabular}}
\label{tools-testing}
\end{table*}

\begin{table*}[!h]
\centering
\caption{Example Tools at the Deployment Stage.}
\scalebox{0.95}{
\begin{tabular}{|c|c|}
\toprule
\multicolumn{2}{|c|}{ \cellcolor{yellow!20} DEPLOYMENT}\\\midrule\midrule
 
Task & Example Tools                       \\ \midrule\midrule

 \multicolumn{2}{|c|}{\bf Q11: Are you monitoring your model?}\\\midrule
 
 Distribution shift detection  &  \makecell{SageMaker model monitor \citeA{nigenda2021amazon} \\ Maximum Mean Discrepency (MMD) \citeA{gretton2012kernel} \\
 Spot-the-diff \citeA{jitkrittum2016interpretable} \\
 Least-Squared density \citeA{rabanser2019failing} \\
 Learned kernel \citeA{liu2020learning} \\
 Learned classifier \citeA{lopez2017revisiting}}
 \\ \midrule
 
 \multicolumn{2}{|c|}{\bf Q12: Do you have mechanisms in place to address data shifts?
}\\\midrule
 Retraining  &  \makecell{Cron jobs \citeA{singh2018dynamic} \\ Workflow orchestrators (Airflow, Prefect) \citeA{singh2019airflow}\\ Platform tools (SageMaker, Azure ML) \citeA{joshi2020amazon,team2016azureml}}
 \\ \midrule
 Failure informed dataset updates  &  \makecell{ Tools needed}
 \\ \midrule

  \multicolumn{2}{|c|}{ \bf Q13: Have you incorporated tools to engender model trust?}\\\midrule
  
  Uncertainty quantification &  \makecell{MC Dropout \citeA{gal2016dropout} \\ Deep Ensembles \citeA{lakshminarayanan2017simple} \\ Conformal prediction \citeA{shafer2008tutorial}} \\ \midrule
  Interpretability/Explainability &  \makecell{Lime \citeA{ribeiro2016should} \\ SHAP \citeA{lundberg2017unified} \\Concept-based \citeA{ghorbani2019towards}} \\ \midrule
  Out-of-distribution detection &  \makecell{Autoencoder reconstruction \citeA{an2015variational,sakurada2014anomaly} \\ Likelihood ratios \citeA{ren2019likelihood} \\ Isolation Forest \citeA{liu2008isolation} \\ Mahalanobis Distance \citeA{lee2018simple}} \\ \midrule
  Fairness &  \makecell{ AIF360 \citeA{bellamy2019ai} \\ Aequitas \citeA{saleiro2018aequitas}}
\\ 
   
\bottomrule
\end{tabular}}
\label{tools-deployment}
\end{table*}

\clearpage

\section{DC-CHECK Worked Examples}
We present two worked examples using DC-CHECK. These serve as examples of how the DC-CHECK questions could be answered and how the checklist could be used more generally.

\begin{itemize}
    \item Example 1: A Dynamic Pipeline for Spatio-Temporal Fire Risk Prediction \citeA{singh2018dynamic}
    \item Example 2: Deep Learning System for the Detection of Diabetic Retinopathy \citeA{beede2020human,gulshan2016development}

\end{itemize}

\clearpage



\section*{Supplementary material}

# DC-CHECK (WORKED EXAMPLE 1)

***Project Description:*** A Dynamic Pipeline for Spatio-Temporal Fire Risk Prediction

***Link:*** https://dl.acm.org/doi/abs/10.1145/3219819.3219913

**Citation:** Singh Walia, B., Hu, Q., Chen, J., Chen, F., Lee, J., Kuo, N., Narang, P., Batts, J., Arnold, G. and Madaio, M., 2018, July. A dynamic pipeline for spatio-temporal fire risk prediction. In *Proceedings of the 24th ACM SIGKDD International Conference on Knowledge Discovery & Data Mining*

| DATA |
|------|
| **Q1: How did you select, collect or curate your dataset?**<br><br>   -  *Have you conducted forensics on the dataset (i.e. provenance)?*<br>   -  *Did you assess the pertinence of the dataset for the task?*<br>   -  *Is your dataset curation a once-off?* |

**Data curation**

We start by acquiring data sets from a variety of sources that contain data we hypothesize (based on prior work) to be relevant in predicting fire risk in non-residential properties.

1. ***Pittsburgh Bureau of Fire:*** Historical fire incident data from 2009-2017, updated on a weekly basis. All fire incident codes are included that have an associated address.
2. ***Allegheny County Office of Property Assessments (OPA):*** property assessment data, updated on a monthly basis, as well as a parcel dataset, which contains information about every parcel in the City of Pittsburgh.
3. ***Pittsburgh Department of Permits, Licenses, and Inspections (PLI):*** records of non-fire inspections and violations (e.g. noise or sanitation violations).
4. ***Residential risk model uses the following additional data:***
- Allegheny County Department of Court Records (DCR): tax lien data, updated monthly
- US Census Bureau: 2012-2016, 5-year estimate American Community Survey (ACS) data

**Data selection**

Datasets are chosen that are likely correlated with fires based on prior work, specifically data about income, occupancy, year built, and the year resident moved in.

**Q2: What data cleaning and/or pre-processing, if any, has been performed?**

   -  *Do you need to "clean" your data?*
   -  *Do you have any missing data?*

1. **Commercial property risk model**

***Joining of data***

Fire incidents were logged at the address level, and thus, to predict fire risk, all the other data first needed to be aggregated across the individual parcels in each address

Join the Allegheny County property assessment data set, with the PLI inspection violations dataset with the non-residential properties (by which we mean Commercial, Industrial, Governmental, or Utility), at the parcel level.

***Data Cleaning:***
- Stripping white spaces from text values
- Drop duplicate columns
- Drop rows with significant (85%) missing values for data
- Aggregate the parcel data at the address level by taking the mean of numeric features and using the most frequent category for categorical features.

**2.  Residential property risk model**

***Joining of data***
Same procedure as above, except that we further aggregated the data to the census block group level by taking the sum for numeric features, which included tax lien, and PLI data, and the most common for categorical features, which included some property data.

***Data Cleaning:***
- Sum as opposed to the mean to capture the total number of violations and inspections in each block, as well as the total amount of unpaid taxes.
- Stripping white space
- Standardizing hyphens
- Standardizing street abbreviations

**BOTH:**
For both models, we then merged the resulting aggregated data frame with the fire incidents dataframe (for both the commercial and residential model).

**Q3: Has data quality been assessed?**

- Completeness, diversity, label quality
- Easy, Hard examples etc

Data quality has not been explicitly assessed, as the data used comes from high-quality governmental data sources with their own checks and balances.

**Q4: Have you considered synthetic data?**

Synthetic data has not been considered as:
- Data privacy is not an issue as it is an internal government deployed project.
- Data access is permitted. Hence, synthetic data is not required.
- Individual datasets are large >100k, hence no need to augment with synthetic data

| TRAINING |
|---|

**Q5: Have you conducted a model architecture and hyperparameter search?**

- *Have you specifically adapted the model for the task?*
- *Is there a potential to incorporate inductive biases?*

**Model architecture:**
Commercial risk model:  Logistic Regression, Ada Boost, Random Forest, XG Boost compared with XG Boost chosen

Residential risk model: Ada Boost, Random Forest, XG Boost compared with Random Forest chosen

**Hyperparameters:**
Grid search is done for the following hyperparameters

- XGBoost commercial risk model, we searched max_depth, min_child_weight, subsample, colsample_by_tree and tuned the rest manually.

- Random Forest residential risk model, we searched n_estimators, max_depth, and max_features using a 1-year validation set.

**Q6: Does the training data match the anticipated use?**

- *Can we anticipate the data the model will be applied on?*
- *Could the model be exposed to data from different distributions/domains?*

**Data matching anticipated use:**
- The model will only be used to provide risk scores for properties in Pittsburgh.

**Distribution/domain shift:**
- There is no expected distribution or domain shift to account for - as we keep the location fixed and will always use the latest data.

**Q7: Are there different data subsets or subgroups of interest?**

- *Are the subgroups identifiable?*
- *Have you assessed fairness/bias introduced by training?*

**Subgroups/subsets of data:**
- There are different property types that can be considered as subgroups/subsets.

**Training based on subgroup/subset:**
- Currently, the models are optimized on average rather than per subgroup.

**Q8: Is the data noisy, either in features or labels?**

**Presence of data noise:**
- There is no specific data noise considered or assessed

**Accounting for data noise:**
- We do not account for data noise or potentially noisy labels, as the data does not come from crowd-sourcing, but rather from official sources.

| TESTING |
|---|

**Q9:  How has the dataset been split for model training and validation?**

- *Are you using a benchmark dataset?*
- *Is the splitting random?*

**Data split**
- Used a walk forward time partition approach, to ensure past events are used to assess future events only
- Training set (6 years of data), validation set for feature selection (1 year of data), and test set (the final 1 year of data).

**Q10: How has the model been evaluated (e.g. metrics & stress tests)?**

- *What metrics are used to assess the model?*
- *Has the model been evaluated beyond average?*
- *Is there a potential to test specific aspects/sub-groups of the model (i.e. stress tests)?*

As the use case of fire prediction, we want to prioritize correctly classifying more of the positive class (i.e. fire) over minimizing false positives (which may result in more inspections, but would be less likely to lead to missed incidents).

**Metrics:**
- Main: kappa and recall
- Secondary: AUC and precision

Cross-validation was performed using the training set.

**Assessment beyond average and/or subgroups:**
- Not automated but "face validity" assessment conducted, where different property types model risk scores were assessed.
- It was found they agree with existing expert knowledge and Bureau of Fire risk reduction efforts

| DEPLOYMENT |
|---|

**Q11:  Are you monitoring your model?**

- *Have you considered dimensionality in the monitoring?*
- *Is there lag in ground truth feedback?*

**Ground truth availability/lag:**
- Ground truth risk scores are not immediately available & would require domain expertise.

**Monitoring:**
- There is no explicit model monitoring, as the model is retrained weekly

(i) Metric being monitored: N/A

(ii) Monitoring method: N/A

**Q13: Do you have mechanisms in place to address data shifts?**

- *Have model updates been considered & when to retrain?*
- *Does your system provide actionable feedback upon failure?*
- *Are you able to characterize the type of shift?*
- *Do system failures inform dataset updates?*

**Model updates**
- Cron job runs every Saturday. Scrapes data sources for the latest dataset, retrains the model, and updates those risk scores on the map and dashboard.

**Failure feedback:**
- Our system does not provide feedback on failures

**Characterizing the type of shift:**
- We do not consider characterizing the type of shift

**Failure informed dataset updates:**
- The dataset is updated weekly, which while not failure informed would help to rectify issues, representing the latest data

**Q14: Have you incorporated tools to engender model trust?**

- *Do you require predictive uncertainty estimates?*
- *Does your ML system have explainability?*
- *Does your system account for OOD inputs?*
- *Do you assess issues of bias and fairness in deployment?*

**Uncertainty estimates:**
- Uncertainty estimates of the risk scores have not been included.

**Explainability:**
- Feature importance scores were computed for samples (Explainability)

**OOD:**
- OOD detection has not been included in the pipeline

**Bias/Fairness:**
- Assessment of model bias or fairness gaps (for example between property type subgroups) have not been considered

# DC-CHECK (WORKED EXAMPLE 2)

***Project Description:*** Deep Learning System for the Detection of Diabetic Retinopathy

***Link:***
[1] https://jamanetwork.com/journals/jama/fullarticle/2588763
[2] https://dl.acm.org/doi/pdf/10.1145/3313831.3376718

| DATA |
|---|
| **Q1: How did you select, collect or curate your dataset?**<br><br>- *Have you conducted forensics on the dataset (i.e. provenance)?*<br>- *Did you assess the pertinence of the dataset for the task?*<br>- *Is your dataset curation a once-off?* |
| **Data curation**<br>Macula-centered retinal fundus images were retrospectively obtained from EyePACS:<br>- United States and<br>- 3 eye hospitals in India (Aravind Eye Hospital, Sankara Nethralaya, and Narayana Nethralaya).<br><br>**Additional validation data collected**<br>● EyePACS screening sites between May 2015 and October 2015. This data set did not overlap with the EyePACS data used in development.<br>● Publicly available Messidor-2 data set, obtained between January 2005 and December 2010 at 3 hospitals in France using a Topcon TRC NW6 nonmydriatic camera and 45° fields of view centered on the fovea.<br><br>Multiple camera types were used: including Centervue DRS, Optovue iCam, Canon CR1/DGi/CR2, and Topcon NW using 45° fields of view. |
| **Q2: What data cleaning and/or pre-processing, if any, has been performed?**<br><br>- *Do you need to "clean" your data?*<br>- *Do you have any missing data?* |
| ● Input images were scale normalized by detecting the circular mask of the fundus image and resizing the diameter of the fundus to be 299 pixels wide. |

- Images for which the circular mask could not be detected were excluded. This corresponded to 117 out of 128,175 on the development set, 17 out of 9,963 in EyePACS-1, and none in Messidor-2.

**Q3: Has data quality been assessed?**

- Completeness, diversity, label quality
- Easy, Hard examples etc

54 graders were used to validate and grade the images. All graders were US-licensed ophthalmologists or ophthalmology trainees in their last year of residency (postgraduate year 4).

A majority vote was taken if the image was considered referable I.e. of good quality, where the threshold is >0.5 agreement across raters.

**Q4: Have you considered synthetic data?**

Synthetic data has not been considered as:
- Data privacy is not an issue as de-identified data is available
- Data access is permitted. Hence, synthetic data is not required.
- Datasets are sufficiently large, hence no need to augment with synthetic data

**TRAINING**

**Q5: Have you conducted a model architecture and hyperparameter search?**

- *Have you specifically adapted the model for the task?*
- *Is there a potential to incorporate inductive biases?*

**Model architecture:**
Ensemble of neural networks used —> Inception v3 selected, pre-trained on ImageNet (adds inductive bias for better convergence)

**Hyperparameters:**
- Not specifically specified, besides the architecture

**Q6: Does the training data match the anticipated use?**

- *Can we anticipate the data the model will be applied on?*
- *Could the model be exposed to data from different distributions/domains?*

**Data matching anticipated use:**
- Yes, the models will be used in the same domain for diabetic retinopathy

**Distribution/domain shift:**
- There is likely a distribution shift if used in different countries or on different quality machines. e.g. Thailand - hence external validation needed

**Q7: Are there different data subsets or subgroups of interest?**

- *Are the subgroups identifiable?*
- *Have you assessed fairness/bias introduced by training?*

**Subgroups/subsets of data:**
- There are different subgroups of diabetic retinopathy: No diabetic retinopathy (45.4%), Mild diabetic retinopathy (25.9%), Moderate diabetic retinopathy (15.1%), Severe diabetic retinopathy (4.5%)

**Training based on subgroup/subset:**
- Currently, the models are optimized on average rather than per subgroup.

**Q8:  Is the data noisy, either in features or labels?**

**Presence of data noise:**
- Noisy labels could be considered as variation in grading

**Accounting for data noise:**
- The quality control process accounts for this by filtering data points - hence no need to account for it during training

---

| TESTING |
|:---:|

**Q9:  How has the dataset been split for model training and validation?**

- *Are you using a benchmark dataset?*
- *Is the splitting random?*

**Data split**
- Random 80-20 train-tuning split of the dataset

**Q10: How has the model been evaluated (e.g. metrics & stress tests)?**

- *What metrics are used to assess the model?*
- *Has the model been evaluated beyond average?*
- *Is there a potential to test specific aspects/sub-groups of the model (i.e. stress tests)?*

**Metrics:**
- Main: area under the receiver operating curve (AUC) generated by plotting sensitivity vs 1- specificity
- Secondary: Sensitivity as a high sensitivity is a prerequisite in a potential screening tool.

**Assessment beyond average and/or subgroups:**
- Additional sensitivity analyses were conducted for several subgroups:
  (1) detecting moderate or worse diabetic retinopathy only;
  (2) detecting severe or worse diabetic retinopathy only;
  (3) detecting referable diabetic macular edema only;
  (4) image quality; and
  (5) referable diabetic retinopathy on 2 data sets, each restricted to mydriatic and non-mydriatic images, respectively

| DEPLOYMENT |
|---|

**Q11: Are you monitoring your model?**

- *Have you considered dimensionality in the monitoring?*
- *Is there lag in ground truth feedback?*

A prospective study was carried out in Thailand & model performance was monitored based on this trial. No automated monitoring pipeline.

**Q13: Do you have mechanisms in place to address data shifts?**

- *Have model updates been considered & when to retrain?*
- *Does your system provide actionable feedback upon failure?*
- *Are you able to characterize the type of shift?*
- *Do system failures inform dataset updates?*

**Model updates**
- Manual or automated model re-training pipelines have not been included

**Failure feedback:**
- The system provides feedback on failures, when the image is of low-quality

**Characterizing the type of shift:**
- We do not consider characterizing the type of shift

**Failure informed dataset updates:**
- The dataset is updated weekly, which while not failure informed would help to rectify issues, representing the latest data

**Q14: Have you incorporated tools to engender model trust?**

- *Do you require predictive uncertainty estimates?*
- *Does your ML system have explainability?*
- *Does your system account for OOD inputs?*
- *Do you assess issues of bias and fairness in deployment?*

**Uncertainty estimates:**
- Uncertainty estimates of the predictions have not been included.

**Explainability:**
- No explicit model explainability
**OOD:**
- Yes, there is a gradeability detection system.
- For patient safety reasons, it only assesses the highest-quality images. If an image has a bit of blur or a dark area, for instance, the system will reject it, even if it could make a strong prediction.

**Bias/Fairness:**
- Assessment of model bias or fairness gaps have not been provided



\bibliographystyleA{unsrt}
\bibliographyA{references}

\end{document}